\documentclass[conference]{IEEEtran}
\IEEEoverridecommandlockouts
\usepackage{cite}
\usepackage{amsmath,amssymb,amsfonts}
\usepackage{algorithmic}
\usepackage{graphicx}
\usepackage{textcomp}
\usepackage{xcolor}
\def\BibTeX{{\rm B\kern-.05em{\sc i\kern-.025em b}\kern-.08em
    T\kern-.1667em\lower.7ex\hbox{E}\kern-.125emX}}
\begin{document}

\title{DINO-BOLDNet: A DINOv3-Guided Multi-Slice Attention Network for T1-to-BOLD Generation \\
}

\author{\IEEEauthorblockN{
Jianwei Wang\textsuperscript{1}\IEEEauthorrefmark{1}, 
Qing Wang\textsuperscript{2}\IEEEauthorrefmark{1},
Menglan Ruan\textsuperscript{3},
Rongjun Ge\textsuperscript{4}\IEEEauthorrefmark{2},
Chunfeng Yang\textsuperscript{1}\IEEEauthorrefmark{2},
Yang Chen\textsuperscript{1}
and Chunming Xie\textsuperscript{2}\IEEEauthorrefmark{2}\\
}

\IEEEauthorblockA{
\textsuperscript{1}School of Computer Science and Engineering\\
Jiangsu Provincial Joint International Research Laboratory of Medical Information Processing\\
Southeast University, Nanjing, China\\
Emails: 220242376@seu.edu.cn,\; chunfeng.yang@seu.edu.cn,\; chenyang.list@seu.edu.cn
}

\IEEEauthorblockA{
\textsuperscript{2}Department of Neurology, Affiliated ZhongDa Hospital, School of Medicine\\
Jiangsu Key Laboratory of Brain Science and Medicine\\
Southeast University, Nanjing, China\\
Emails: drwangqing0601@163.com,\; chmxie@163.com
}
\IEEEauthorblockA{
\textsuperscript{3}School of Software Engineering\\
Jiangsu Provincial Joint International Research Laboratory of Medical Information Processing\\
Southeast University, Nanjing, China\\
Emails: 230250010@seu.edu.cn
}
\IEEEauthorblockA{
\textsuperscript{4}School of Instrument Science and Engineering\\
Southeast University, Nanjing, China\\
Emails: rongjun\_ge@seu.edu.cn
}
\IEEEauthorblockA{
\IEEEauthorrefmark{1}These authors contributed equally to this work.\\
\IEEEauthorrefmark{2}Corresponding authors: Rongjun Ge, Chunfeng Yang, Chunming Xie
}
}
\maketitle

\begin{abstract}
Generating BOLD images from T1w images offers a promising solution for recovering missing BOLD information and enabling downstream tasks when BOLD images are corrupted or unavailable. Motivated by this, we propose DINO-BOLDNet, a DINOv3-guided multi-slice attention framework that integrates a frozen self-supervised DINOv3 encoder with a lightweight trainable decoder. The model uses DINOv3 to extract within-slice structural representations, and a separate slice-attention module to fuse contextual information across neighboring slices. A multi-scale generation decoder then restores fine-grained functional contrast, while a DINO-based perceptual loss encourages structural and textural consistency between predictions and ground-truth BOLD in the transformer feature space.  Experiments on a clinical dataset of 248 subjects show that DINO-BOLDNet surpasses a conditional GAN baseline in both PSNR and MS-SSIM.  To our knowledge, this is the first framework capable of generating mean BOLD images directly from T1w images, highlighting the potential of self-supervised transformer guidance for structural-to-functional mapping. 

\end{abstract}


\section{Introduction}
Structural T1-weighted (T1w) images provide stable and high-resolution anatomical information and are widely available in clinical and research datasets. In contrast, blood-oxygen-level-dependent (BOLD) functional images capture dynamic neural activity and functional organization that cannot be inferred from structural imaging alone, yet they are often missing or unusable in real-world settings\cite{luh2000comparison}. This imbalance between reliable structural imaging and frequently unavailable functional imaging motivates generating missing BOLD information directly from the more stable T1w modality. 

Vision transformers (ViTs) are highly effective at capturing rich global representations, making them well suited for tasks that require broad anatomical context\cite{hatamizadeh2022unetr}. Pretrained ViTs also transfer strongly to medical imaging applications, avoiding the need to train large CNN-based models such as U-Net\cite{ronneberger2015u} from scratch. As a self-supervised ViT, DINOv3\cite{simeoni2025dinov3} learns stable and expressive global structural features from large-scale pretraining, and these representations have demonstrated strong generalization to medical domains. Such powerful global encoding makes DINOv3 particularly suitable for structural-to-functional generation, where mapping anatomical structure to functional contrast involves distributed relationships that must be learned from limited data. 

Motivated by these factors, We propose DINO-BOLDNet, a DINOv3-based multi-slice attention framework that generates BOLD images directly from T1w images. The model integrates transformer-derived structural embeddings into a generative architecture that preserves global semantic correspondence while restoring fine-grained functional contrast. Experiments on a clinical dataset of 248 subjects demonstrate that the generated mean BOLD images are anatomically consistent and functionally meaningful, providing a promising solution for recovering functional information when BOLD data are missing or unusable.  

In summary, our contributions are shown as follows:

\begin{itemize}
    \item We present a framework for  generating a subject-specific mean BOLD image directly from T1w image, addressing the widespread issue of missing or unusable BOLD image in clinical and research cohorts and enabling functional preprocessing even when BOLD images are unavailable.
    \item We propose a DINOv3-based cross-modal generation architecture that integrates transformer representations, slice-attention fusion, and multi-scale decoding, allowing the model to extract fine-grained structural cues from T1 and learn a stable structural-to-functional mapping under limited training data.
    \item We demonstrate that the generated mean BOLD images are anatomically consistent and functionally meaningful, achieving strong peak signal-to-noise ratio (PSNR) and multi-scale structural similarity (MS-SSIM) performance. Experiments on a 248-subject clinical dataset show that our generated mean BOLD images closely approximate the real BOLD images.
\end{itemize}

\begin{figure}
    \centering
    \includegraphics[width=1\linewidth]{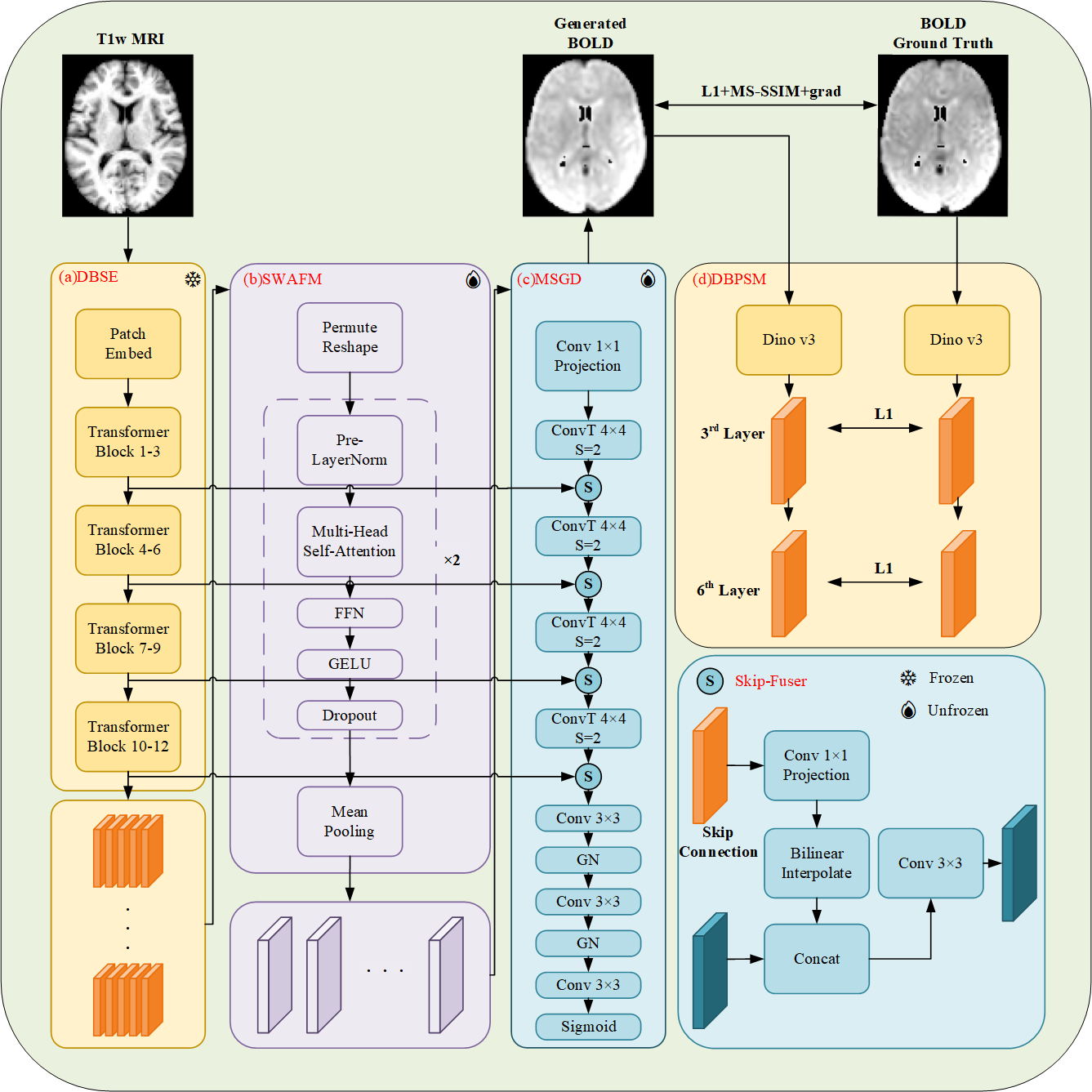}
    \caption{\textbf{
    Overview of the proposed DINOv3-guided T1-to-BOLD generation framework.}
 \textit{(a) DBSE: DINO-based structural encoder; (b) SWAFM: slice-wise attention fusion module; (c) MSGD: multi-scale Generation Decoder; (d) DBPSM: DINO-based perceptual supervision module.}
}
    \label{fig:placeholder}    
\end{figure}  

\section{Methods}

As shown in Fig.~\ref{fig:placeholder}, DINO-BOLDNet consists of four key modules: DINO-based structural encoder, slice-wise attention fusion module, multi-scale generation decoder, and DINO-based perceptual supervision module.Four modules are detailed in the following sections.

\subsection{DINO-based Structural Encoder}

Let $X \in \mathbb{R}^{H \times W \times Z}$ denote a preprocessed T1w image and $Y \in \mathbb{R}^{H \times W \times Z}$ denote the corresponding mean BOLD image. Our goal is to learn a nonlinear mapping
$f_\theta : X \rightarrow \hat{Y}$,
such that the generated image $\hat{Y}$ approximates $Y$ voxel-wise.

To accommodate the input format of DINOv3, we reformulate the 3D generation task in a slice-based manner. For each axial index $z$, we extract a multi-slice window 
$W_z = \{ X_{z-(K-1)/2}, \dots, X_z, \dots, X_{z+(K-1)/2} \}$, 
where odd $K$ denotes the number of slices in the window, ensuring that $X_z$ is centered. Boundary slices are padded when needed and tracked via a binary mask. Each slice is replicated across channels and resampled to $224 \times 224$, yielding an input tensor of shape $B \times K \times 3 \times 224 \times 224$.

As illustrated in Fig.~\ref{fig:placeholder}(a), a pretrained DINOv3 ViT-B/16 encoder is used as a frozen structural feature extractor. For each slice in the window, patch embedding followed by $L$ transformer blocks produces a token matrix $T^{(L)}_k \in \mathbb{R}^{196 \times D}$, where $196=14\times14$ corresponds to the patch grid. Stacking tokens from all $K$ slices yields a 3D token tensor $T^{(L)} \in \mathbb{R}^{K \times 196 \times D}$.

To obtain hierarchical structural representations, we additionally extract intermediate outputs from selected transformer blocks $l \in \{3,6,9,12\}$. These outputs, originally represented as $S^{(l)}_k \in \mathbb{R}^{196 \times D}$ for each slice, are stacked across slices to form $S^{(l)} \in \mathbb{R}^{K \times 196 \times D}$. In our framework, these intermediate representations serve as skip connections (SC), providing multi-level anatomical cues that will later be incorporated into the decoder. All DINOv3 parameters remain frozen during training to preserve the pretrained feature space and improve data efficiency.

\subsection{Slice-wise Attention Fusion Module}

Although each slice is encoded independently by DINOv3, the functional contrast at a given spatial location often depends on anatomical context distributed across adjacent slices. To capture such cross-slice dependencies, we introduce a slice-wise attention fusion module, as illustrated in Fig.~\ref{fig:placeholder}(b).

For each spatial patch location, the tokens extracted from the $K$ slices are stacked into a tensor $T \in \mathbb{R}^{K \times 196 \times D}$. Multi-head self-attention is then applied along the slice dimension, enabling each slice to attend to its neighbors and aggregate contextual information. The resulting fused tokens $\tilde{T}$ thus encode 3D-aware structural cues.  
A similar slice-wise attention operation is applied to each skip connection $S^{(l)}$ (from transformer blocks $l \in \{3,6,9,12\}$), producing fused skip representations $\tilde{S}^{(l)}$ that preserve multi-level anatomical context.

\subsection{Multi-scale Generation Decoder}

As illustrated in Fig.~\ref{fig:placeholder}(c), the decoder reconstructs a dense BOLD slice from the fused representation produced by the slice-wise attention module. The input to the decoder is the fused main-branch feature $\tilde{T}$, which is first reshaped into a coarse spatial map $F^{(0)} \in \mathbb{R}^{D \times 14 \times 14}$. This serves as the initial decoding resolution.

At each decoding stage, the upsampled features are fused with the corresponding slice-attended skip connections $\tilde{S}^{(l)}$ obtained from transformer blocks $l \in \{3,6,9,12\}$. Each skip tensor is reshaped to a $14 \times 14$ grid and resized via bilinear interpolation to match the spatial resolution of the current decoding layer. The decoder concatenates the resized skip feature with the upsampled representation and applies a $3 \times 3$ convolution, group normalization (GN), and GELU activation to refine spatial structure and local detail.

This multi-scale fusion strategy progressively combines global context from $\tilde{T}$ with fine anatomical cues carried by $\tilde{S}^{(l)}$, enabling accurate recovery of boundaries and functional contrast. A final $3 \times 3$ convolution maps the refined features to a single-channel prediction at $224 \times 224$, which is then resampled and unpadded using the mask $M$ to restore the original in-plane resolution. Repeating this process across all axial indices yields the final output $\hat{Y}$.

\subsection{DINO-based Perceptual Supervision Module}

As shown in Fig.~\ref{fig:placeholder}(d), this module provides high-level structural guidance for T1-to-BOLD generation by enforcing feature-level consistency between the synthesized prediction and the ground-truth BOLD image. While pixel-wise losses supervise local intensity patterns, perceptual alignment encourages anatomical and textural correspondence within the pretrained DINOv3 feature space.

To compute perceptual supervision, both the predicted slice $\hat{Y}$ and the target slice $Y$ are fed through the frozen DINOv3 encoder. Let $\phi_l(\cdot)$ denote features extracted from transformer block $l$. The perceptual loss is defined as:
\begin{equation}
\label{eq:perc}
\mathcal{L}_{\mathrm{perc}}
= \sum_{l \in \{3,6\}}
\left\| \phi_l(\hat{Y}) - \phi_l(Y) \right\|_1 .
\end{equation}

\paragraph{Masked L1 loss}
This term enforces voxel-wise intensity fidelity within the valid region, ensuring that the generated BOLD values match the target at a local, pixel-wise level:
\begin{equation}
\label{eq:l1}
\mathcal{L}_{\mathrm{L1}}
= \left\| M \odot (\hat{Y} - Y) \right\|_1 ,
\end{equation}
where $M$ is the valid-region mask and $\odot$ denotes element-wise multiplication.

\paragraph{MS-SSIM loss}
This term promotes structural similarity by encouraging the generator to preserve contrast patterns and spatial relationships that are perceptually meaningful in neuroimaging:
\begin{equation}
\label{eq:msssim}
\mathcal{L}_{\mathrm{MS\text{-}SSIM}}
= 1 - \mathrm{MS\text{-}SSIM}(M \odot \hat{Y},\, M \odot Y).
\end{equation}

\paragraph{Gradient loss}
This loss enhances boundary sharpness and encourages accurate recovery of local transitions, which is important for preserving functional interfaces and regional contrast variations:
\begin{equation}
\label{eq:grad}
\mathcal{L}_{\mathrm{grad}}
= \left\| \nabla_x \hat{Y} - \nabla_x Y \right\|_1
+ \left\| \nabla_y \hat{Y} - \nabla_y Y \right\|_1 .
\end{equation}

\paragraph{Total loss}
The final training objective is a weighted sum of all components:
\begin{equation}
\label{eq:total}
\mathcal{L}
= \lambda_1 \mathcal{L}_{\mathrm{L1}}
+ \lambda_s \mathcal{L}_{\mathrm{MS\text{-}SSIM}}
+ \lambda_g \mathcal{L}_{\mathrm{grad}}
+ \lambda_p \mathcal{L}_{\mathrm{perc}},
\end{equation}
where $\lambda_1$, $\lambda_s$, $\lambda_g$, and $\lambda_p$ balance the contributions of pixel-wise accuracy, structural similarity, edge preservation, and perceptual alignment.

\section{Experiments and Results
}
\subsection{Dataset and Preprocessing          }

We used structural and functional image data from 248 participants collected at Zhongda Hospital, Southeast University\cite{zhang2021altered}, spanning the Alzheimer's disease spectrum: Alzheimer's disease (AD, $n=30$), mild cognitive impairment (MCI, $n=70$), subjective cognitive decline (SCD, $n=56$), and cognitively normal controls (CN, $n=92$). All subjects underwent high-resolution 3D structural T1-weighted imaging using a 3D-MPRAGE sequence, and BOLD was acquired using a gradient-echo echo-planar imaging (GRE-EPI) sequence on the same scanner.

To ensure consistent preprocessing across modalities, we adopted DeepPrep\cite{ren2025deepprep} as a unified pipeline for both T1w and BOLD images. For T1w images, DeepPrep\cite{ren2025deepprep} performed bias-field correction, skull stripping, and deep-learning-based anatomical alignment to a standardized template space. For BOLD data, the pipeline included head motion correction, slice-timing correction, susceptibility distortion correction, and rigid followed by nonlinear registration to the MNI152 template.

After preprocessing the BOLD time series, we discarded the first 10 time-point images  to mitigate magnetization instability and transient adaptation effects. The subject-specific mean BOLD image $Y$ was then computed by averaging the remaining time points along the temporal dimension. Finally, both T1w images and mean BOLD images were resampled to a common MNI152 space with matched voxel resolution, yielding spatially aligned images for subsequent modeling. In our framework, the preprocessed T1w image serves as the input modality, and the corresponding mean BOLD image is the regression target.
\begin{figure}
    \centering
    \includegraphics[width=1\linewidth]{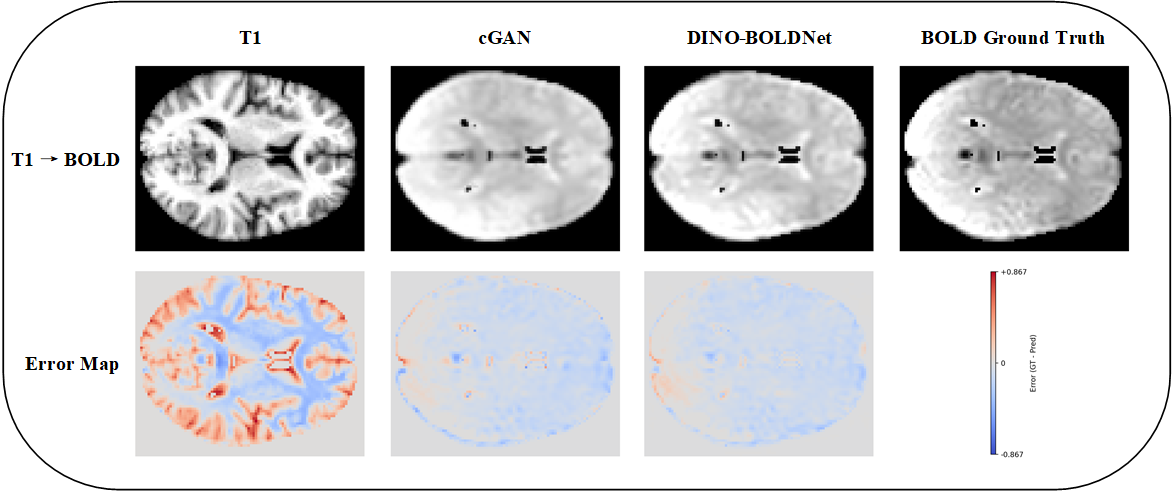}
   \caption{Visual comparison of T1-to-BOLD generation. Top: input T1, cGAN output, DINO-BOLDNet output, and ground-truth BOLD. Bottom: error maps (GT -- Pred), where red/blue denote positive/negative residuals. DINO-BOLDNet shows clearer anatomical detail and lower generation error.}

    \label{fig:comparison}

\end{figure}
\subsection{Experimental Setup 
}
 All experiments were implemented in PyTorch and executed on an RTX 3090 GPU. DINO-BOLDNet was trained for 100 epochs using the AdamW optimizer with a cosine annealing schedule and a batch size of 32. The DINOv3 encoder remained frozen throughout training, while the slice-attention fuser and multi-scale decoder were fully trainable\cite{loshchilov2017decoupled}.

To assess generation quality, we adopt two widely used evaluation metrics: PSNR, which measures voxel-wise fidelity, and MS-SSIM\cite{wang2003multiscale}, which captures perceptual and structural consistency and is particularly suitable for low-contrast neuroimaging data. All metrics are reported in the original spatial resolution of the generated mean BOLD images.

 The complete training and architectural hyperparameters of DINO-BOLDNet are summarized in Table~\ref{tab:hyper}.

\begin{table}[t]
\centering
\caption{Training and architectural hyperparameters of the proposed DINO-BOLDNet.}
\label{tab:hyper}
\setlength{\tabcolsep}{10pt}
\begin{tabular}{lc}
\hline
\textbf{Hyperparameter} & \textbf{Configuration}\\
\hline
\multicolumn{2}{l}{\textbf{Training Settings}} \\
Optimizer & AdamW \\
Initial Learning Rate (LR)& $2\times10^{-4}$ \\
Weight Decay & $1\times10^{-4}$ \\
LR Scheduler & Cosine Annealing \\
Minimum Learning Rate & $1\times10^{-6}$ \\
Batch Size & 32 \\
Max Epochs & 100 \\
Train/Val Split & 80\% / 20\% \\
\hline
\multicolumn{2}{l}{\textbf{Input Format}} \\
Input Resolution & $224 \times 224$ \\
Slice Window Size ($K$) & 5\\
\hline
\multicolumn{2}{l}{\textbf{Model Architecture}} \\
Token Dimension ($D$) & 768 \\
Slice Attention Heads & 4 \\
Attention Layers & 2 \\
Dropout Rate & 0.1 \\
Decoder Base Channels & 512 \\
\hline
\multicolumn{2}{l}{\textbf{Loss Weights}} \\
$\lambda_{\mathrm{L1}}$ & 1.0 \\
$\lambda_{\mathrm{SSIM}}$ & 0.5 \\
$\lambda_{\mathrm{Grad}}$ & 0.1 \\
$\lambda_{\mathrm{Perc}}$ & 0.05 \\
\hline
\end{tabular}
\end{table}

\begin{table}[t]
\centering
\caption{Comparison between cGAN and DINO-BOLDNet on original-resolution generation.}
\label{tab:comparison}
\setlength{\tabcolsep}{8pt} 
\begin{tabular}{lcc}
\hline
Method & PSNR & MS-SSIM \\
\hline
cGAN & 30.41 & 0.9588 \\
Ours & \textbf{31.10} & \textbf{0.9613} \\
\hline
\end{tabular}
\end{table}

\begin{table}[t]
\centering
\caption{Ablation study on architectural and loss components.}
\label{tab:ablation}
\setlength{\tabcolsep}{5pt}
\begin{tabular}{ccc|cc}
\hline
SA& SC& Loss& PSNR& MS-SSIM\\
\hline
$\times$ & $\times$ & Full            & 29.61 & 0.9008 \\
$\times$ & \checkmark & Full          & 30.52 & 0.8955 \\
\checkmark & \checkmark & L1           & \textbf{30.93}& 0.9062 \\
\checkmark & \checkmark & L1 + MS-SSIM& 30.75 & 0.9043 \\
\checkmark & \checkmark & L1 + MS-SSIM + Grad& 30.80 & 0.9052 \\
\checkmark & \checkmark & Full& 30.86& \textbf{0.9062} \\
\hline
\end{tabular}
\end{table}

\subsection{Quantitative Comparison }

Table \ref{tab:comparison} reports the generation performance of DINO-BOLDNet and a cGAN baseline at the original spatial resolution. DINO-BOLDNet achieves higher PSNR and MS-SSIM values, with PSNR increasing from 30.41 dB to 31.10 dB and MS-SSIM increasing from 0.9588 to 0.9613. A higher PSNR indicates more accurate recovery of voxel-wise intensity, while a higher MS-SSIM reflects better preservation of structural and contrast information across scales. These results suggest that DINO-BOLDNet reproduces both the intensity distribution and the anatomical structure of the mean BOLD image more faithfully. A qualitative comparison is presented in Fig. \ref{fig:comparison}.

\subsection{Ablation Study}

To assess the contribution of each component in DINO-BOLDNet, we conducted ablation experiments focusing on the slice-wise attention fusion module (SWAFM), the skip connections (SC), and different loss combinations. The results are summarized in Table\ref{tab:ablation}.

Configurations without SWAFM show a clear reduction in PSNR, suggesting that incorporating cross-slice contextual information plays an important role in capturing structural cues that influence BOLD contrast. Excluding skip connections also leads to a decline in performance, especially in MS-SSIM, reflecting the value of multi-scale feature pathways for maintaining anatomical coherence during generation.

We further examined how different loss formulations affect reconstruction quality. Using only L1 loss results in the highest PSNR but leads to noticeably smoother outputs. Adding MS-SSIM and gradient losses improves structural sharpness, while the full loss formulation provides the most balanced outcome, yielding competitive voxel-level fidelity together with better preservation of anatomical structure and contrast.

\section{Conclusion}

This work presents DINO-BOLDNet, a DINOv3-guided multi-slice attention framework for generating mean BOLD images directly from structural T1w images. By combining pretrained DINOv3 representations with cross-slice contextual fusion, the model captures structural cues that are informative for predicting functional contrast. Quantitative and qualitative results show that DINO-BOLDNet produces higher PSNR and MS-SSIM scores and yields clearer anatomical detail compared with the baseline. Ablation experiments further highlight the contributions of slice-wise attention, skip connections, and DINOv3-based perceptual supervision.

Overall, DINO-BOLDNet provides an effective approach for recovering functional information in cases where BOLD images are corrupted or unavailable.

\bibliographystyle{IEEEtran}
\bibliography{cite}

\begin{thebibliography}{1}
\providecommand{\url}[1]{#1}
\csname url@samestyle\endcsname
\providecommand{\newblock}{\relax}
\providecommand{\bibinfo}[2]{#2}
\providecommand{\BIBentrySTDinterwordspacing}{\spaceskip=0pt\relax}
\providecommand{\BIBentryALTinterwordstretchfactor}{4}
\providecommand{\BIBentryALTinterwordspacing}{\spaceskip=\fontdimen2\font plus
\BIBentryALTinterwordstretchfactor\fontdimen3\font minus \fontdimen4\font\relax}
\providecommand{\BIBforeignlanguage}[2]{{%
\expandafter\ifx\csname l@#1\endcsname\relax
\typeout{** WARNING: IEEEtran.bst: No hyphenation pattern has been}%
\typeout{** loaded for the language `#1'. Using the pattern for}%
\typeout{** the default language instead.}%
\else
\language=\csname l@#1\endcsname
\fi
#2}}
\providecommand{\BIBdecl}{\relax}
\BIBdecl

\bibitem{luh2000comparison}
W.-M. Luh, E.~C. Wong, P.~A. Bandettini, B.~D. Ward, and J.~S. Hyde, ``Comparison of simultaneously measured perfusion and bold signal increases during brain activation with t1-based tissue identification,'' \emph{Magnetic Resonance in Medicine: An Official Journal of the International Society for Magnetic Resonance in Medicine}, vol.~44, no.~1, pp. 137--143, 2000.

\bibitem{hatamizadeh2022unetr}
A.~Hatamizadeh, Y.~Tang, V.~Nath, D.~Yang, A.~Myronenko, B.~Landman, H.~R. Roth, and D.~Xu, ``Unetr: Transformers for 3d medical image segmentation,'' in \emph{Proceedings of the IEEE/CVF winter conference on applications of computer vision}, 2022, pp. 574--584.

\bibitem{ronneberger2015u}
O.~Ronneberger, P.~Fischer, and T.~Brox, ``U-net: Convolutional networks for biomedical image segmentation,'' in \emph{International Conference on Medical image computing and computer-assisted intervention}.\hskip 1em plus 0.5em minus 0.4em\relax Springer, 2015, pp. 234--241.

\bibitem{simeoni2025dinov3}
O.~Sim{\'e}oni, H.~V. Vo, M.~Seitzer, F.~Baldassarre, M.~Oquab, C.~Jose, V.~Khalidov, M.~Szafraniec, S.~Yi, M.~Ramamonjisoa \emph{et~al.}, ``Dinov3,'' \emph{arXiv preprint arXiv:2508.10104}, 2025.

\bibitem{zhang2021altered}
Q.~Zhang, Q.~Wang, C.~He, D.~Fan, Y.~Zhu, F.~Zang, C.~Tan, S.~Zhang, H.~Shu, Z.~Zhang \emph{et~al.}, ``Altered regional cerebral blood flow and brain function across the alzheimer's disease spectrum: a potential biomarker,'' \emph{Frontiers in Aging Neuroscience}, vol.~13, p. 630382, 2021.

\bibitem{ren2025deepprep}
J.~Ren, N.~An, C.~Lin, Y.~Zhang, Z.~Sun, W.~Zhang, S.~Li, N.~Guo, W.~Cui, Q.~Hu \emph{et~al.}, ``Deepprep: an accelerated, scalable and robust pipeline for neuroimaging preprocessing empowered by deep learning,'' \emph{Nature Methods}, pp. 1--4, 2025.

\bibitem{loshchilov2017decoupled}
I.~Loshchilov and F.~Hutter, ``Decoupled weight decay regularization,'' \emph{arXiv preprint arXiv:1711.05101}, 2017.

\bibitem{wang2003multiscale}
Z.~Wang, E.~P. Simoncelli, and A.~C. Bovik, ``Multiscale structural similarity for image quality assessment,'' in \emph{The thrity-seventh asilomar conference on signals, systems \& computers, 2003}, vol.~2.\hskip 1em plus 0.5em minus 0.4em\relax Ieee, 2003, pp. 1398--1402.

\end{thebibliography}

\end{document}